\documentclass{article}
\usepackage{spconf,amsmath,graphicx,hyperref,booktabs, multirow, amsfonts}

\title{Retrieval-Augmented Generation for \\ Electrocardiogram-Language Models}
\name{
\begin{tabular}{c}
Xiaoyu Song$^{*1}$, William Han$^{*1}$\thanks{* Equal contribution}, Tony Chen$^{2}$, Chaojing Duan$^{3}$,\\
Michael A. Rosenberg$^{4}$, Emerson Liu$^{3}$, Ding Zhao$^{1}$ \thanks{© 2026 IEEE. Personal use of this material is permitted. Permission from IEEE must be obtained for all other uses, in any current or future media, including reprinting/republishing this material for advertising or promotional purposes, creating new collective works, for resale or redistribution to servers or lists, or reuse of any copyrighted component of this work in other works.}
\end{tabular}
}
\address{$^{1}$Carnegie Mellon University, $^{2}$Columbia University, \\$^{3}$Allegheny Health Network, $^{4}$University of Colorado}

\usepackage{color}

\begin{document}
%
\maketitle
\begin{abstract}
Interest in generative Electrocardiogram-Language Models (ELMs) is growing, as they can produce textual responses conditioned on ECG signals and textual queries. Unlike traditional classifiers that output label probabilities, ELMs are more versatile, supporting domain-specific tasks (e.g., waveform analysis, diagnosis, prognosis) as well as general tasks (e.g., open-ended questions, dialogue). Retrieval-Augmented Generation (RAG), widely used in Large Language Models (LLMs) to ground LLM outputs in retrieved knowledge, helps reduce hallucinations and improve natural language generation (NLG). However, despite its promise, no open-source implementation or systematic study of RAG pipeline design for ELMs currently exists. To address this gap, we present the first open-source RAG pipeline for ELMs, along with baselines and ablation studies for NLG. Experiments on three public datasets show that ELMs with RAG consistently improves performance over non-RAG baselines and highlights key ELM design considerations. Our code is available at: https://github.com/willxxy/ECG-Bench.
\end{abstract}

\section{Introduction}
Cardiovascular diseases (CVDs) are the leading global cause of death, responsible for about 18 million deaths annually \cite{worldhealthorganization_2024_cardiovascular}. Electrocardiograms (ECGs) are noninvasive, widely available, and central to early CVD detection. However, ECGs still require expert clinicians for accurate interpretations. This is a growing challenge given rising screening demands and a nationwide shortage of cardiac specialists, particularly in under-served regions \cite{10.1001/jamainternmed.2024.2270}. To address this gap, deep learning has been applied to automate ECG-based diagnosis centering around classification tasks \cite{qiu-etal-2023-transfer}. More recently, large language model (LLM)-based generative approaches, termed Electrocardiogram-Language Models (ELMs), have emerged. Unlike traditional classification models that provide only probability scores for fixed labels, ELMs incorporate knowledge from large-scale pretraining on internet data and can be adapted to output not just CVD labels but also textual explanations that justify the diagnostic decisions \cite{zhao2024ecgchatlargeecglanguagemodel}. This expanded capability reduces the burden on cardiac electrophysiologists, who are often tasked with examining vast volumes of ECG data, and enables clinicians without specialized training to interact with sophisticated ECG interpretation tools.

\begin{figure}
    \centering
    \includegraphics[width=1\linewidth]{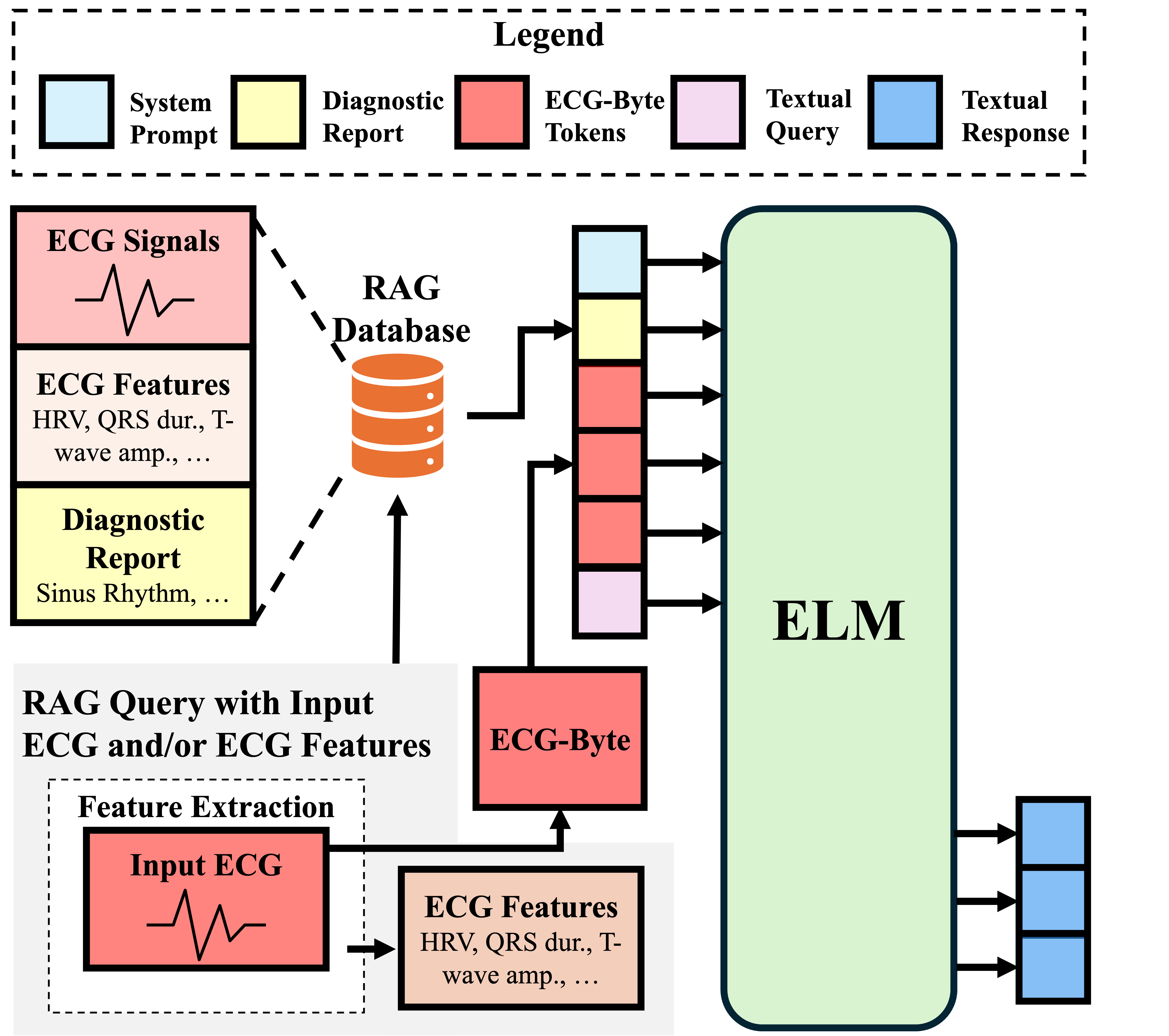}
    \caption{Our RAG pipeline operates as follows: given an input ECG, we optionally extract features and query a RAG database of ECG signals, features, and diagnostic reports. We retrieve the top-k similar diagnostic reports, construct a prompt (system prompt + retrieved diagnostic reports + ECG tokens + query), and use it to condition the ELM to generate the response.}
    \label{fig:main}
\end{figure}

In natural language processing (NLP), Retrieval-Augmented Generation (RAG) improves large language model (LLM) outputs by retrieving relevant external documents. This reduces hallucinations and enables more accurate, context-aware generation beyond the model’s parametric memory \cite{yu2024rankragunifyingcontextranking}. 

Applying RAG to ECG deep learning has largely focused on classification and retrieval tasks \cite{pmlr-v225-qiu23a}. ECG-Chat \cite{zhao2024ecgchatlargeecglanguagemodel} is among the first to extend RAG to free-form NLG for ELMs, but it (1) does not provide an open-source RAG pipeline and (2) does not analyze how RAG design choices affect performance. Another related work, Q-Heart \cite{pham2025qheartecgquestionanswering}, incorporates RAG content during instruction tuning to update input prompts dynamically. However, this work currently lacks an open-source implementation, hindering direct comparison, and it treats RAG primarily as a tool for boosting question answering without exploring design variations. Similarly, \cite{pmlr-v225-yu23b} applies RAG with LLMs for ECG diagnosis, but their approach (1) lacks an open-source implementation, (2) encodes handcrafted ECG features (e.g., heart rate variability, QRS duration) as text instead of directly using ECG signals, and (3) does not evaluate free-form NLG, despite producing textual explanations. 

To address these gaps, we present the first open-source RAG framework for training and inference in ELMs tailored to NLG. Additionally, we comprehensively evaluate the performance of including RAG during NLG across three publicly available datasets and by ablating various components in the pipeline. In summary, our contributions are the following:\\
1. To the best of our knowledge, we are the first to develop an open source framework for RAG in NLG for ELMs.\\ 
2. We conduct baselines on three public datasets and demonstrate strong performance gains when utilizing RAG across different ELM architectures.\\
3. To inform researchers about RAG pipeline design trade-offs, we conduct a comprehensive ablation study, varying training–inference combinations with RAG, the number of retrieved items k, the placement of retrieved content, and the effect of noise injection in the retrieved content.

Our open-source implementation allows for direct comparisons, and the comprehensive ablation study on integrating RAG into ELMs provides valuable insights for future researchers.

\section{Methods}
\subsection{Datasets and Preprocessing}
This study leverages adapted versions of the MIMIC-IV-ECG~\cite{johnson_2023_mimiciv} and PTB-XL~\cite{wagner_ptb-xl_2020} datasets for NLG tasks. We incorporate the ECG-Chat Instruct dataset curated by \cite{zhao2024ecgchatlargeecglanguagemodel}, which includes single- and multi-turn conversational data paired with a corresponding ECG. In addition, we utilize the ECG-QA dataset \cite{oh2023ecgqacomprehensivequestionanswering}, which contains ChatGPT-generated, clinically relevant question-answer pairs derived from both MIMIC-IV-ECG and PTB-XL. 

We apply a unified preprocessing pipeline to all ECG signals, following ECG-Byte \cite{han2024ecgbytetokenizerendtoendgenerative}. Signals are first standardized to the PTB-XL lead configuration [I, II, III, aVL, aVR, aVF, V1–V6]. Powerline noise at 50/60 Hz is removed with bidirectional notch filters (Q=30), and clinically relevant components are preserved using a fourth-order Butterworth bandpass filter (0.5–100 Hz). Baseline drift is corrected with a bidirectional high-pass filter (0.05 Hz). For denoising, we apply wavelet decomposition (Daubechies-6, level 4) with soft thresholding based on the median absolute deviation of detail coefficients. All signals are resampled to 250 Hz and segmented into non-overlapping 5-second windows.

\begin{table*}[!htp]\centering
\caption{Mean baseline comparisons over 3 random seeds across three datasets.}\label{tab:main}
\scriptsize
\begin{tabular}{llcccccc}\toprule
Dataset &Method &BLEU-4 $\uparrow$ (\%) &ROUGE-L $\uparrow$ (\%) &METEOR $\uparrow$ (\%)&BERTScore F1 $\uparrow$ (\%)&Accuracy $\uparrow$ (\%) \\\midrule
\multirow{2}{*}{ECG-Chat Instruct \cite{zhao2024ecgchatlargeecglanguagemodel}} &ECG-Byte \cite{han2024ecgbytetokenizerendtoendgenerative} & 22.85 ± 0.18 & 66.82 ± 0.17 & 58.97 ± 0.16 & 96.24 ± 0.02 & 9.65 ± 0.29 \\
&ECG-Byte with RAG & \textbf{38.10 ± 0.05} & \textbf{75.61 ± 0.05} & \textbf{69.85 ± 0.02} & \textbf{97.49 ± 0.01} & \textbf{18.27 ± 0.05}\\
\midrule
\multirow{2}{*}{ECG-QA PTB-XL \cite{oh2023ecgqacomprehensivequestionanswering}} &ECG-Byte \cite{han2024ecgbytetokenizerendtoendgenerative} & 21.07 ± 0.05 & \textbf{48.47 ± 0.02} & \textbf{33.66 ± 0.07} & 96.37 ± 0.00 & 37.03 ± 0.04 \\
&ECG-Byte with RAG & \textbf{21.46 ± 0.13} & 48.00 ± 0.03 & 32.20 ± 0.02 & \textbf{96.44 ± 0.01} & \textbf{38.30 ± 0.06 }\\
\midrule
\multirow{2}{*}{ECG-QA MIMIC-IV ECG \cite{oh2023ecgqacomprehensivequestionanswering}} &ECG-Byte \cite{han2024ecgbytetokenizerendtoendgenerative} & 15.52 ± 0.03 & 39.75 ± 0.11 & 26.62 ± 0.08 & 95.28 ± 0.01 & 27.21 ± 0.12  \\
&ECG-Byte with RAG & \textbf{49.07 ± 0.14}& \textbf{76.32 ± 0.00} & \textbf{59.51 ± 0.06} & \textbf{97.98 ± 0.00} & \textbf{57.77 ± 0.12}\\
\bottomrule
\end{tabular}
\end{table*}

\subsection{Retrieval-Augmented Generation Database}
We curate a domain-specific RAG database composed of ECG signal segments, derived features, and corresponding diagnostic reports. This database is designed to support multimodal similarity search using both ECG signals and ECG features.

We extract features from the time, frequency, and time–\\frequency domains for each lead of an ECG signal. The time-domain features include maximum, minimum, heart rate, heart rate variability, QRS duration, T-wave amplitude, ST deviation, average absolute difference, and root mean square difference. The frequency-domain features include total power, peak frequency power, dominant frequency, and spectral centroid. The time–frequency features include wavelet coefficient approximation and detail levels (up to level 5). These features are used to construct the RAG database. The same set of features are derived from the input ECG and computed on the fly during training and inference to serve as the query to the RAG database.


We use the FAISS library’s \cite{douze2024faiss} IndexIVFFlat structure for efficient vector indexing. For each ECG signal and/or feature query, we search the corresponding index and retrieve the top-$k$ nearest neighbors based on L2 distance. Each ECG signal and feature index is linked to its associated diagnostic report, allowing us to also retrieve the top-$k$ nearest diagnostic reports. In our study, we query the RAG database utilizing both ECG features and signals unless specified otherwise. During training or inference we query the RAG database and insert the retrieved content (i.e., top-k diagnostic reports) in the system prompt on the fly, as shown in Figure~\ref{fig:main}.


\subsection{Electrocardiogram-Language Model}
\begin{figure}
    \centering
    \includegraphics[width=1\linewidth]{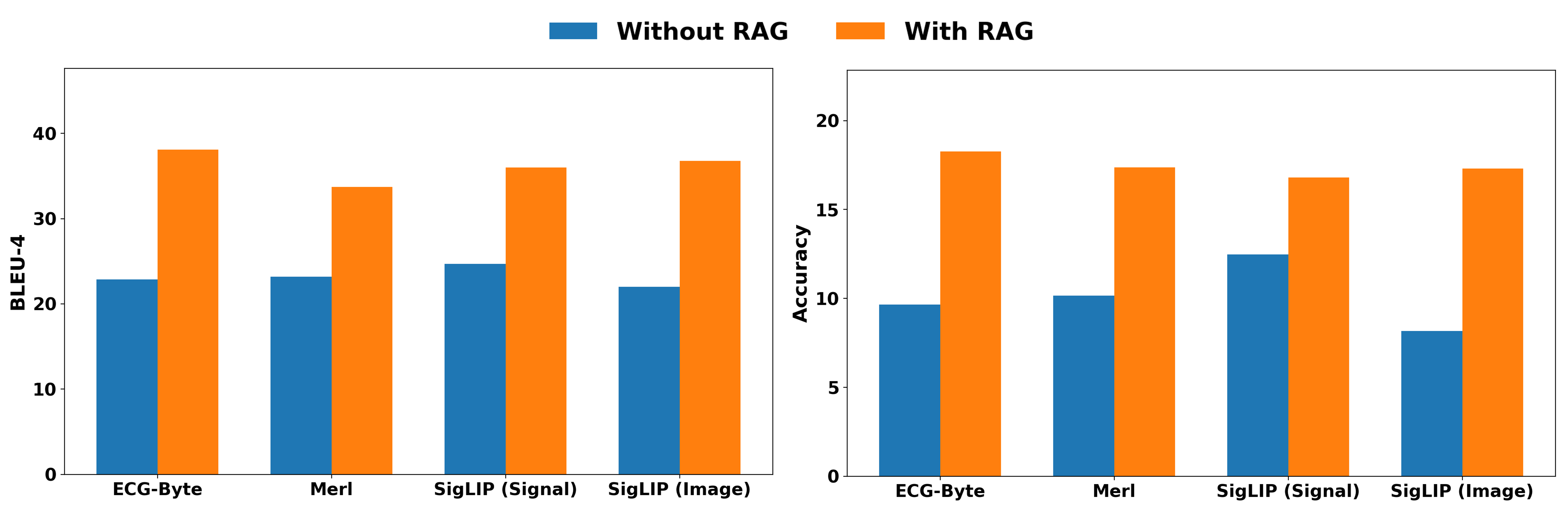}
    \caption{Our RAG pipeline demonstrates flexibility across multiple ELM architectures while consistently improving BLEU-4 and accuracy.}
    \label{fig:encoder}
\end{figure}

Several variants of ELMs have been explored \cite{han2025signalimagesymbolicexploring}, but we mainly adopt ECG-Byte \cite{han2024ecgbytetokenizerendtoendgenerative} due to its simplicity and low computational overhead. Unlike neural network encoder–based ELMs, ECG-Byte does not compress the ECG signal through a neural network encoder. Instead, it applies the Byte Pair Encoding (BPE) algorithm \cite{sennrich2016neuralmachinetranslationrare} to create ECG tokens. ECG-Byte has shown strong performance compared with encoder-based approaches \cite{han2025signalimagesymbolicexploring}, making it our primary method for all experiments unless otherwise specified. Building on ECG-Byte, we employ the Llama-3.2-1B-Instruct checkpoint \cite{grattafiori2024llama3herdmodels} via the HuggingFace Transformers API \cite{wolf2020huggingfaces}, using default hyperparameters except for the addition of new ECG signal tokens to the LLM’s embedding layer as described in \cite{han2024ecgbytetokenizerendtoendgenerative}. To demonstrate that our RAG pipeline is not limited to ECG-Byte, we also report baselines with other ELM architectures in Figure~\ref{fig:encoder}.

\subsection{Learning Objective}
Following previous work \cite{han2025signalimagesymbolicexploring}, we consider an autoregressive objective that is compatible with general conversational input formats. Each input sequence is represented as $Y = (y_1, y_2, \dots, y_T)$,
constructed by concatenating a system prompt \( q_{\text{sys}} \), retrieved information \( q_{\text{rag}} \), a signal token sequence \( X_{\text{ID}} \), the initial user query \( q_1 \) and assistant response \( s_1 \), and a sequence of query-response pairs \( q_2, s_2, \dots, q_n, s_n \).

To ensure that learning focuses exclusively on generating the assistant responses, we define a target labeling function $\ell: \{1, \dots, T\} \to \mathcal{V} \cup \{-100\}$, where \( \mathcal{V} \) is the model’s vocabulary. The labeling function is \[
\ell(t) = 
\begin{cases}
y_t, & \begin{aligned}
       &\text{if } y_t \in s_i \\
       &\text{or end-of-turn token,}
       \end{aligned} \\
-100, & \text{otherwise}
\end{cases}
\]
The value \(-100\) is used to mask out non-target tokens (\( q_{\text{sys}} \), \( q_{\text{rag}} \), \( X_{\text{ID}} \), \( q_n \)) during loss computation.

Letting $\mathcal{T} = \{ t \in \{1, \dots, T\} : \ell(t) \neq -100 \}$, the loss function is defined as
\[
\mathcal{L}(\theta) = - \sum_{t \in \mathcal{T}} \log p_\theta(y_t \mid y_{<t}).
\]

This formulation ensures that only assistant responses \( S = \{ s_1, \dots, s_n \} \) and end-of-turn tokens contribute to the training loss. Retrieved content is treated purely as conditioning context and is excluded from supervision.

\begin{table}[!htp]\centering
\caption{Ablation study on different combinations of training with RAG and inferencing with RAG.}\label{tab:rag_train}
\resizebox{\columnwidth}{!}{%
\begin{tabular}{lccccc}\toprule
Method &Training w/ RAG &Inference w/ RAG &BLEU-4 $\uparrow$ (\%) &Accuracy $\uparrow$ (\%) \\
\midrule
ECG-Byte \cite{han2024ecgbytetokenizerendtoendgenerative} & & &22.85 ± 0.18 & 9.65 ± 0.29 \\
\midrule
\multirow{4}{*}{ECG-Byte with RAG} 
&\checkmark & & 20.08 ± 0.11 &7.17 ± 0.03 \\
\cmidrule(r){2-5} 
& &\checkmark & 32.33 ± 0.10 & 14.96 ± 0.12 \\
\cmidrule(r){2-5} 
&\checkmark &\checkmark & \textbf{38.10 ± 0.05} & \textbf{18.27 ± 0.05} \\
\bottomrule
\end{tabular}
}
\end{table}

\section{Experiments}
\subsection{Experimental Settings}
\label{exp:set}
We utilize the Adam optimizer \cite{kingma2017adam} with a learning rate of 1e-4 and weight decay of 1e-2. For all experiments, we train the ELMs for 1 epoch with a batch size of 2 over 400,000 randomly sampled ECGs. We inference on a separate test set of size 20,000 instances. For the input length of the ELM, we pad/truncate inputs to a fixed size of 1024. We apply Low-Rank Adaptation (LoRA) \cite{hu2021loralowrankadaptationlarge} with $\text{rank} = 16$. All experiments were completed on NVIDIA RTX A6000 48 GB GPUs.

We describe the experimental settings for each ELM considered. Following ECG-Byte \cite{han2024ecgbytetokenizerendtoendgenerative}, we train a tokenizer with 3,500 merges on 10-second unsegmented ECGs. We implement three neural network encoder-based ELMs from prior work \cite{han2025signalimagesymbolicexploring}: Merl, SigLIP (Signal), and SigLIP (Image). We pretrain Merl on 800,000 ECG instances for 50 epochs using a Res-Net101 backbone. For SigLIP-based ELMs, we use the siglip-base-patch16-224 checkpoint from HuggingFace. The two SigLIP variants differ only in input: SigLIP (Signal) uses stacked signal representations, while SigLIP (Image) uses plotted ECG images \cite{han2025signalimagesymbolicexploring}. We follow previous work \cite{han2025signalimagesymbolicexploring} and adopt a LLaVA-based approach \cite{liu2023visualinstructiontuning}, directly applying SigLIP without additional finetuning. For all encoder-based ELMs, we freeze the encoder during LLM training and add a projection layer jointly trained with the LLM. Lastly, the three neural network encoder-based ELMs use the same Llama-3.2-1B-Instruct checkpoint as the LLM.

\section{Results}

\subsection{Baselines}
We present NLG results with and without RAG using ECG-Byte, averaged across three random seeds and datasets, in Table~\ref{tab:main}. Nearly all metrics show improvements when incorporating our RAG pipeline. To demonstrate that our pipeline is not limited to ECG-Byte, we also report averaged results with and without RAG integration for other ELM architectures (as described in section~\ref{exp:set}) on the ECG-Instruct dataset in Figure~\ref{fig:encoder}. These results also show clear performance gains in both BLEU-4 and accuracy when using our RAG pipeline.

\subsection{Ablation Study}
We ablate RAG usage across training and inference, top-$k$ retrieval size, RAG placement in the input prompt, and noise injection. Unless specified otherwise, experiments use the ECG-Chat Instruct dataset with defaults of k=1, RAG in both training and inference and system-prompt insertion. All results are averaged over 3 random seeds.

\begin{table}[!htp]\centering
\caption{Ablation study on varying the number of top k retrieved content.}\label{tab:k}
\resizebox{0.5\columnwidth}{!}{%
\begin{tabular}{lccc}\toprule
k  &BLEU-4 $\uparrow$ (\%)&Accuracy $\uparrow$ (\%)\\\midrule

1 & \textbf{38.10 ± 0.05} & \textbf{18.27 ± 0.05} \\
5& 37.99 ± 0.04  & 18.07 ± 0.11 \\
10 & 36.91 ± 0.09  & 17.20 ± 0.14 \\
\bottomrule
\end{tabular}
}
\end{table}

\textbf{RAG During Training/Inference}
In Table~\ref{tab:rag_train}, we present results when introducing RAG during training and/or inference. Although RAG is often used only at inference time to ground the LLM's output, prior work has introduced RAG during pretraining or finetuning to help the LLM learn the format of the inserted content \cite{gao2024retrievalaugmentedgenerationlargelanguage}. We consider three settings in Table~\ref{tab:rag_train}: 1) using RAG only during training, 2) using RAG only during inference, 3) using RAG during both training and inference.
We find that using RAG during both training and inference yields the best performance, while using RAG only during training and not at inference performs worse than the baseline in Table~\ref{tab:main} (without RAG). Using RAG only during inference improves over the baseline but remains below the performance achieved when RAG is used during both training and inference.

\textbf{Varying Top-k}
We evaluate model performance when altering the number of retrieved items k (Table~\ref{tab:k}). Overall, the differences are minor, with k=1 yielding slightly higher scores than larger k values. While increasing k provides access to more retrieved information, prior work \cite{yu2024rankragunifyingcontextranking} indicates that this does not necessarily translate into better performance. As k grows, the additional content may contain irrelevant or redundant information, effectively acting as noise.

\begin{table}[!htp]\centering
\caption{Investigating the effect of injecting the RAG content in the system prompt or user query.}\label{tab:rag_loc}
\scriptsize
\resizebox{0.75\columnwidth}{!}{%
\begin{tabular}{lcc}\toprule
RAG Location &BLEU-4 $\uparrow$ (\%)&Accuracy $\uparrow$ (\%)\\
\midrule
System prompt &\textbf{38.08 ± 0.10} & 18.11 ± 0.10\\
User query & 38.03 ± 0.03 & \textbf{18.17 ± 0.23} \\
\bottomrule
\end{tabular}
}
\end{table}

\textbf{RAG Location}
Where to place RAG content in the prompt remains an open and underexplored question in NLP \cite{park2025emulatingretrievalaugmentedgeneration}. In Table~\ref{tab:rag_loc}, we find that inserting the retrieved content into either the system prompt or the user query yields comparable results. This similarity may arise because, in both locations, the retrieved information becomes available to the model before it generates its response, shaping the contextual foundation of the interaction. However, subtle differences in placement could influence how the model interprets the retrieved content; either as background knowledge (system prompt) or as part of the user’s intent (user query), which we leave to future work for further exploration.

\begin{table}[!htp]\centering
\caption{The effect of injecting noise into RAG.}\label{tab:noise}
\scriptsize
\begin{tabular}{lccc}\toprule
Method &BLEU-4 $\uparrow$ (\%) &Accuracy $\uparrow$ (\%) \\\midrule
ECG-Byte \cite{han2024ecgbytetokenizerendtoendgenerative} & 22.85 ± 0.18  & 9.65 ± 0.29 \\
ECG-Byte with RAG (Noise) & 23.48 ± 0.05 & 8.19 ± 0.05\\
ECG-Byte with RAG & \textbf{38.10 ± 0.05}  & \textbf{18.27 ± 0.05}\\
\bottomrule
\end{tabular}
\end{table}

\textbf{Injecting noise into RAG}
To test the effect of retrieval quality, we inject noise into the RAG content by replacing diagnostic reports with ``--------------------------'' (Table~\ref{tab:noise}). ECG-Byte with and without noisy RAG show similar performance, while incorporating correct reports yields clear gains, highlighting the importance of accurate retrieval.

\section{Discussion and Conclusions}
We present the first open-source RAG pipeline for ELMs, showing consistent improvements in NLG across multiple datasets and architectures. We also highlight four main findings from our ablation studies: (1) using RAG during both training and inference yields the highest performance, (2) retrieving fewer items (e.g., top-1) marginally outperforms larger retrievals, (3) RAG content placement (system prompt vs. user query) produces similar outcomes, and (4) accurate retrieval is necessary for performance improvements. These insights provide practical guidance for the design of RAG-enabled ELMs and establish a reproducible foundation for future work.\\
\textbf{Acknowledgments} This work is done in collaboration with the Mario Lemieux Center for Heart Rhythm Care at Allegheny General Hospital.

\bibliographystyle{IEEEbib}
\bibliography{refs}
\end{document}